%% file: neurips_2020.tex
\documentclass{article}

     \usepackage[preprint,nonatbib]{neurips_2020}

\usepackage[utf8]{inputenc} 
\usepackage[T1]{fontenc}    
\usepackage{hyperref}       
\usepackage{url}            
\usepackage{booktabs}       
\usepackage{amsfonts}       
\usepackage{nicefrac}       
\usepackage{microtype}      
\usepackage{graphicx}
\usepackage{paralist}
\usepackage{xspace}
\usepackage{xcolor}
\usepackage{caption}
\usepackage{subcaption}
\usepackage[normalem]{ulem}
\usepackage{pbox}
\usepackage[numbers]{natbib}
\newcommand{\sknote}[1]{{\color{magenta}[SK: #1]}}
\newcommand{\anote}[1]{{\color{orange}[AD: #1]}}
\newcommand{\knote}[1]{{\color{red}[KK: #1]}}
\newcommand{\jnote}[1]{{\color{blue}[JC: #1]}}
\newcommand{\ignore}[1]{{[\sout{#1]}}}
\renewcommand{\sknote}[1]{}
\renewcommand{\anote}[1]{}
\renewcommand{\knote}[1]{}
\renewcommand{\jnote}[1]{}
\newcommand{\name}{\textsc{Resonator}\xspace}
\newcommand{\orchestrator}{\text{orchestrator}\xspace}
\newcommand{\ie}{{\em i.e., \/}}
\newcommand{\eg}{{\em e.g., \/}}
\newcommand{\etc}{{\em etc. \/}}

\newcommand{\Scut}[1] {}

\title{Spatial Sharing of GPU for Autotuning DNN models}

\author{
    Aditya Dhakal\\
    University of California, Riverside\\
    \texttt{adhak001@ucr.edu}
    \And
    Junguk Cho\\
    Hewlett Packard Labs\\
    \texttt{junguk.cho@hpe.com}
    \And
    Sameer G. Kulkarni\\
    Indian Institute of Technology, Gandhinagar\\
    \texttt{sameergk@iit.ac.in}
    \And
    K. K. Ramakrishnan\\
    University of California, Riverside\\
    \texttt{kk@cs.ucr.edu}
    \And
    Puneet Sharma\\
    Hewlett Packard Labs\\
    \texttt{puneet.sharma@hpe.com}
}

\begin{document}
\maketitle

\begin{abstract}
\input{sections/abstract.tex}

\end{abstract}

\vspace{-3mm}\section{Introduction}\vspace{-3mm}
\input{sections/introduction}

\vspace{-2mm}\section{Background}\label{sec:background}\vspace{-1mm}
\input{sections/motivation}

\vspace{-2mm}\section{GPU Multiplexing Design}\vspace{-3mm}
\input{sections/design}

\vspace{-3mm}\section{Evaluation}\vspace{-3mm}
\input{sections/evaluation}
\vspace{-2mm}\section{Related Work}\vspace{-2mm}
\input{sections/related}
\vspace{-3mm}\section{Summary}\vspace{-3mm}
\input{sections/summary}

\bibliographystyle{plain}
\bibliography{neurips_2020}

\end{document}

%% file: sections/abstract.tex
GPUs are used for training, inference, and tuning the machine learning models.  
However, Deep Neural Network (DNN) models vary widely in their ability to exploit the full power of high-performance GPUs.
\Scut{With their significant cost and power consumption, 
it is desirable to find ways to utilize GPUs more efficiently.}
Spatial sharing of GPU enables multiplexing several applications on the GPU and can improve utilization of the GPU, thus improving throughput and lowering latency.
DNN models given just the right amount of GPU resources can still provide low inference latency, just as much as dedicating all of the GPU for their inference task. 
An approach to improve DNN inference performance is hardware-specific tuning of the DNN model. Autotuning frameworks find the optimal low-level implementation for a certain target device based on the trained machine learning model, thus reducing the DNN's inference latency and increasing inference throughput. 
We observe an inter-dependency between the tuned model and its inference latency. A DNN model tuned with specific GPU resources provides the best inference latency when inferred with close to the same amount of GPU resources. However, a model tuned with 
the maximum amount of the GPU's resources has poorer inference latency once the GPU resources are limited for inference. On the other hand, a model tuned with an appropriate amount of GPU resources still achieves good inference latency across a wide range of GPU resource availability. 
We explore the underlying causes that impact the tuning of a model at different amounts of GPU resources. 
We present a number of techniques to maximize resource utilization and improve tuning performance. We enable controlled spatial sharing of GPU to multiplex several tuning applications on the GPU. We scale the tuning server instances and shard the tuning model across multiple client instances for concurrent tuning of different operators of a model, achieving better GPU multiplexing. 
With our improvements, we decrease DNN autotuning time by upto 75\% and increase throughput by a factor of 5.  

%% file: sections/introduction.tex

Deep Learning (DL) powered inference use cases (e.g., industrial monitoring, autonomous driving) increasingly common. The high cost of performing the inference (both in hardware cost and power consumption) makes it critical to optimize the inference performance and as well to improve the user's Quality of Experience (QoE).
For example, Facebook needs to serve tens of trillions of inference requests in real time, which demands enormous amount of compute servers~\cite{fc.ml.2018}. Thus, there is significant demand for low-latency, high throughput, \emph{ but still} at high accuracy. Deep\Scut{\knote{why do you say distributed???}} Neural Network (DNN) models for inference services.
Several methods have been proposed to increase the accuracy and to speedup the DNN inference~\cite{DBLP:journals/corr/MolchanovTKAK16,isscc_2016_chen_eyeriss,jouppi2017datacenter}. 
GPUs are typically utilized to provide the necessary acceleration to achieve low-latency inference. 
Another complementary activity is to \emph{tune} the DNN model to run efficiently on a specific hardware platform that the inference will be executed on, to find an optimal model configuration that can best utilize the hardware towards achieving very low-latency inference. Despite these efforts, tuned DNN models that are available today fail to utilize the power of current GPUs efficiently~\cite{jeon2019analysis}. 
GPU utilization by a DNN is further limited by several other factors such as data transfer to the GPU~\cite{dhakal2019netml}, memory access, irregular computation~\cite{hill2017deftnn}, \etc We have also observed that some DNNs have lower GPU utilization due to the smaller computation requirement for some of their layers.\looseness-1 

The objective of our work is two-fold: i) to provide optimally tuned DNN models for low-latency inference; and ii) to improve the tuning efficiency (increasing tuning throughput by reducing tuning time and multiplexing tuning instances) and GPU utilization.
Towards this goal, we design mechanisms to effectively multiplex and concurrently execute multiple 
inference applications on a GPU.
We leverage NVIDIA Multi-Process Service (MPS) and extend it by 
providing an appropriate amount of GPU resources (a GPU\%, \ie restricting the GPU resources provided for inference and for the tuning service by specifying the number of GPU Streaming Multiprocessors (SMs) that an application can use) to the DNN models to achieve low latency inference, while effectively freeing up the GPU to support running more tuning or inference applications concurrently. Therefore, multiplexing several applications on a GPU by spatially sharing the GPU can greatly increase its utilization.

We specifically explore the impact of tuning a model with constrained GPU resources (\ie when the model is tuned with a specified GPU\%), on its inference latency when the inference is performed with different GPU resource limits. As a motivating example we tuned a ResNet-18 model in TVM twice, once by setting GPU\% to 100 and also by setting it to 25. We see in Fig.~\ref{fig:tuning_vs_inference_impact} that tuning a model with the maximum amount of GPU resources (100\% GPU) provides a tuned model that works best only when most or entire GPU is dedicated for inference. However, that same model performs worse than a model tuned with a lower GPU\% when both perform inference with fewer GPU resources (e.g., 25\% GPU or less). In fact, a model tuned at 'just the right' GPU percentage performs better for a wider range of GPU\% during inference. On the other hand, a model tuned at the extremes (too much or too little GPU resource) performs poorly during inference with a GPU \% not matched to that extreme tuning \%.
We present our observations about inter-dependency between a model tuned at a particular GPU\% and its corresponding inference latency in \S~\ref{sec:spatial_sharing_tuning} and experimentally examine and illustrate the reasons for such inter-dependency. 
Further, we devise techniques to spatially share the GPU to decrease tuning latency \emph{and} increase tuning throughput.\vspace{-1mm} 
\Scut{However, sharing a GPU to run multiple applications concurrently can cause interference among them during tuning. This interference can cause the tuning process to report the wrong metrics, therefore resulting in a non-optimally tuned model. We show the effect of uncontrolled GPU sharing on tuning, and that our method to spatially share the GPU carefully allows tuning to arrive at a 'properly' tuned model \S\ref{sec:multiplexing_on_gpu}.} 
\vspace{-0.2cm}

%% file: sections/motivation.tex
\begin{figure}[h]
\vspace{-4mm}
\centering
        \begin{subfigure}[b]{0.32\textwidth}
        \centering
        \includegraphics[width=\textwidth]{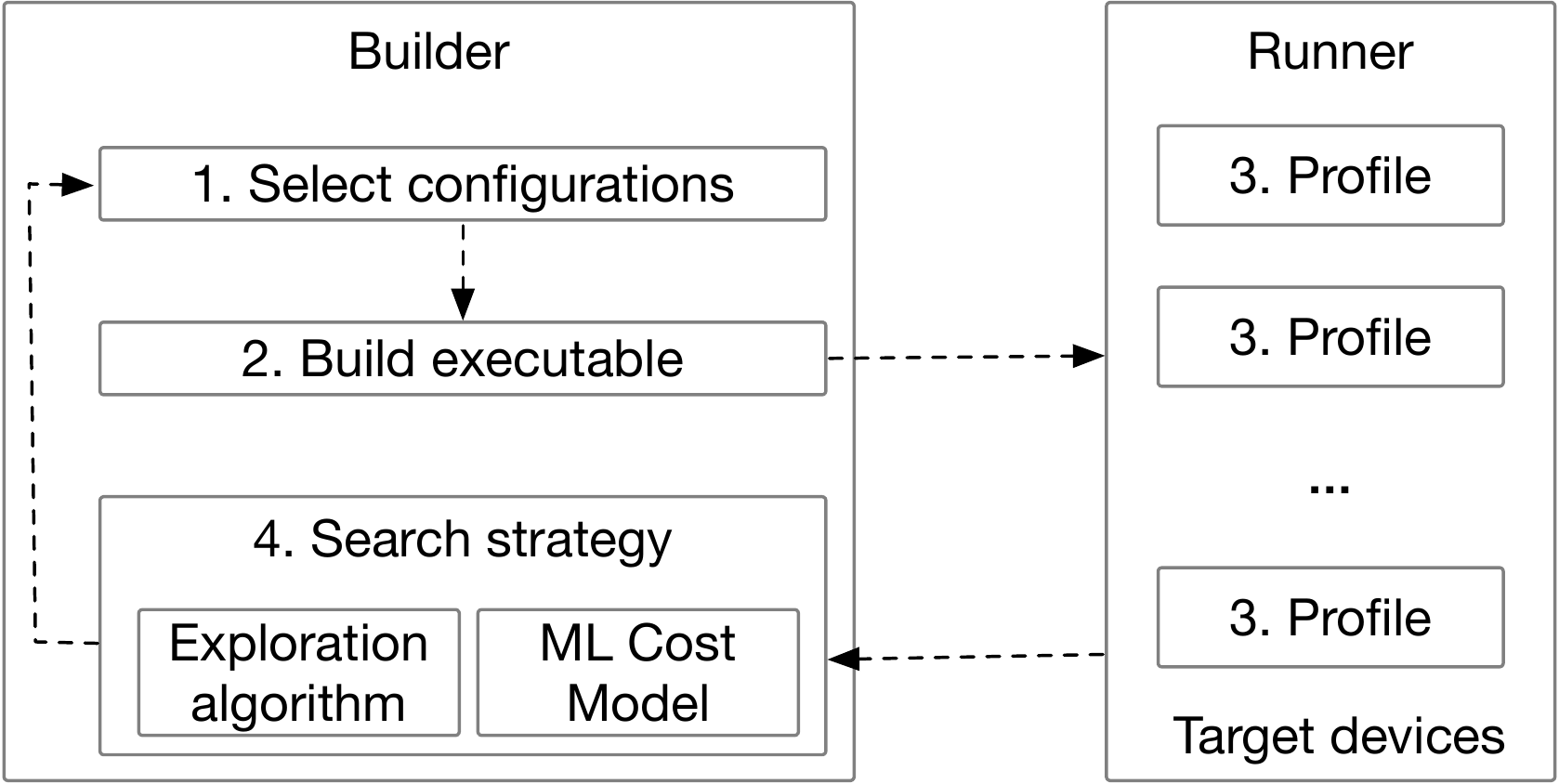}
        \vspace{-4mm}
        \caption{Generic Autotuning Systems}
        \label{fig:tvm.autotuning.steps}
        \end{subfigure}%
        \begin{subfigure}[b]{0.32\textwidth}        
        \centering
        \includegraphics[width=\textwidth]{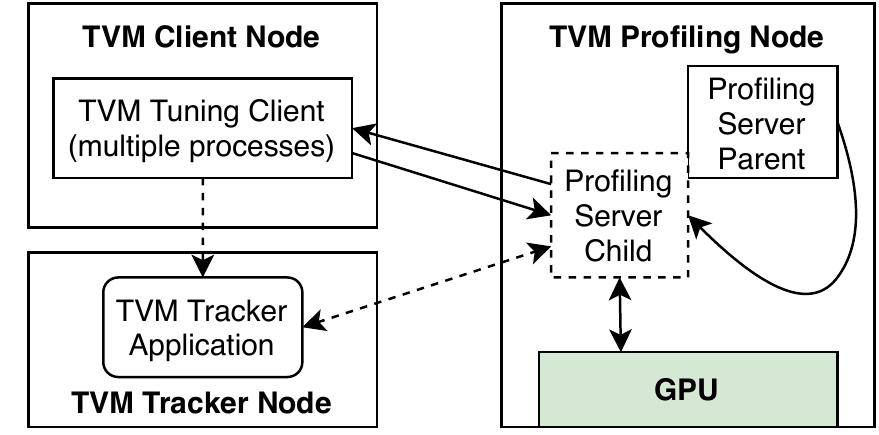}
        \vspace{-4mm}
        \caption{TVM Autotuning Setup}
        \label{fig:original_tvm}
        \end{subfigure}%
        \begin{subfigure}[b]{0.30\textwidth}
            \includegraphics[width=\textwidth]{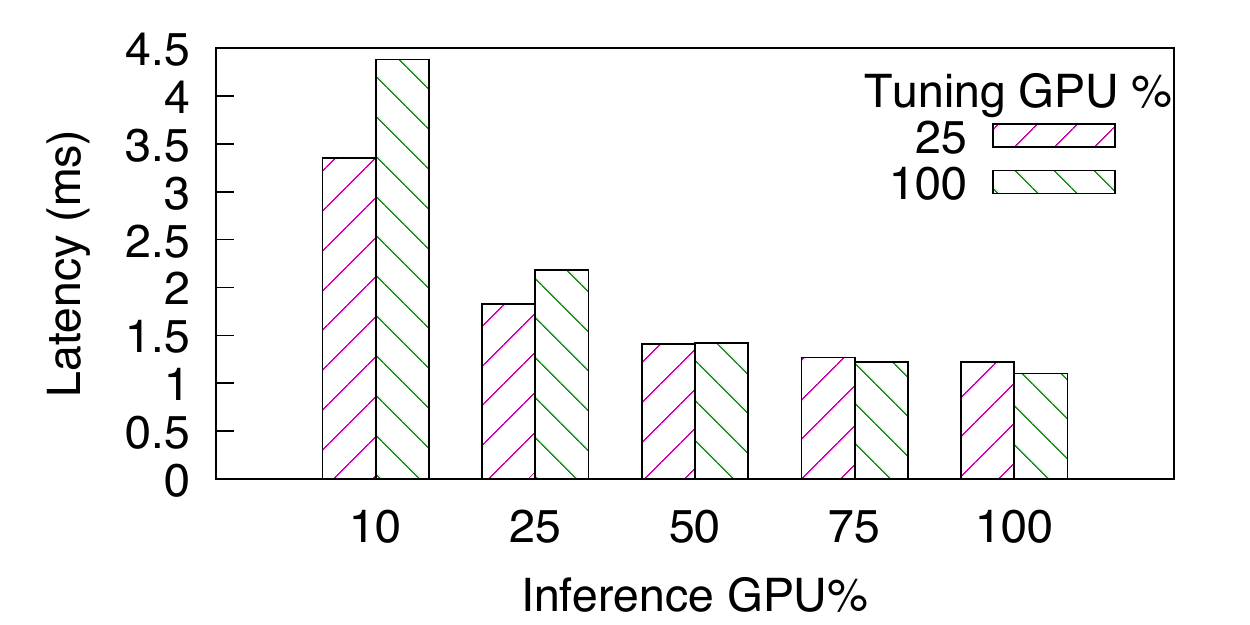}
            \vspace{-4mm}
            \caption{Tuning\% - Infer. latency}
            \label{fig:tuning_vs_inference_impact}
        \end{subfigure}%
        \vspace{-3mm}
\caption{Autotuning systems; GPU \% at tuning and inference interdependency - inference latency}
\label{fig:back.figures}
\vspace{-0.6cm}
\end{figure}
\subsection{DNN Autotuning Systems}\vspace{-3mm}
\textit{Autotuning} or the automated performance optimization of DNN models creates optimized low-level implementations of DNN operators (\eg  2D convolution, fully connected layers, pooling) 
for a specific target hardware (e.g., GPU, FPGA, CPU, etc) to improve inference performance without the need for manual tuning human experts. 
Specifically, autotuning finds optimal configurations for loop tiles and ordering, caching, and loop unrolling to reduce the memory access cost, maximize parallelism (e.g., CUDA threading), and leverage novel hardware primitives (e.g., Tensor Cores) wherever possible in the target hardware. The high-level procedure of autotuning is to have a large number of iterations of sequential evaluations of different DNN model configurations before finding an optimal one.
Fig.~\ref{fig:tvm.autotuning.steps} shows a generic autotuning
procedure to optimize inference performance on target devices. It consists of four stages:
\begin{inparaenum}[(1)]
\item \textbf{Select configurations:} selects a batch of candidate configurations in a search space based on the \textit{Search strategy} stage. In case no initial training data exists, this stage picks random candidate configurations.
\item \textbf{Build:} generates executable files based on this batch of candidates.
\item \textbf{Profile:} runs executable files and measures the execution time on the target device(s). Note that the autotuning procedure has to run on a specific target device that we want to optimize the model for.
\item \textbf{Search strategy:} selects the next promising candidates in a search space and consists of an exploration algorithm and a Machine Learning (ML) cost model.
The exploration algorithm (e.g., simulated annealing~\cite{sa}, and reinforcement learning-based
search~\cite{ahn2019reinforcement, chameleon}) is used to reduce the search space to select the next configurations in the search space with the ML cost model (e.g., XGboost~\cite{chen2016xgboost}, Graph Neural Network~\cite{tomczak2019simulating}). Since every iteration has the order of billions of possible configurations, it is very important to reduce the search space to make it possible to use in real world DL workloads.
The ML cost model is trained based on selected configurations and their corresponding execution times measured in the \textit{Profile} stage and used to predict the execution times of configurations in the exploration algorithm without hardware measurements.
The Builder in fig.~\ref{fig:tvm.autotuning.steps} performs stages 1,2, and 4, and Runner does stage 3.

\end{inparaenum}
{\bf TVM AutoTuning System}
is a popular DNN autotuning application~\cite{chen2018tvm} whose  
architecture facilitates tuning DNNs across multiple compute nodes (Fig.~\ref{fig:original_tvm}). 
\newline\textbf{TVM RPC Client} functions  as the \textit{Builder} process of the generic platform. It has: i) Schedule Explorer, searches for and proposes new configurations that provide more optimal DNN operators. The proposed configurations are first compiled by the target compiler (\eg NVCC for NVIDIA GPUs), and then transferred to the TVM RPC server for profiling; 
ii) Cost Model, for reducing the search space and overall tuning time. TVM also has an early stopping module that halts tuning if a newer configuration generated by the explorer is worse than previously profiled configurations. 
\newline\textbf{TVM RPC Server} is the generic \textit{Runner} process running 
 on the server compute node with the target hardware (\eg GPU). It 
gets the compiled code 
from the TVM RPC Client, executes them on the target hardware and reports the result back to the TVM RPC client; 
\newline\textbf{TVM Tracker} acts as a broker that coordinates and arbitrates the tuning process between the TVM RPC client and server components and additionally provides authentication and other security measures necessary to restrict/control the client-server interaction. 
\vspace{-2mm}\subsection{Impact of spatially sharing GPU with fixed percentage on tuning and inference}\label{sec:spatial_sharing_tuning}\vspace{-2mm}

Effectively utilizing the GPU by multiplexing tasks running concurrently is challenging. The default NVIDIA GPU runtime environment executes one application at a time, even if there are enough GPU resources to allow multiple tasks. Applications share the GPU temporally by executing their GPU kernels in a fixed time quantum provided by GPU scheduler. This unfortunately increases the overall latency for all concurrently running applications. The CUDA Multi-Process System (MPS)~\cite{nvidia.mps} has been introduced to make concurrent execution of applications possible by spatially sharing the GPU to reduce idling of GPU resources. But, with default CUDA MPS, a compute heavy application can impact resources for other applications running concurrently, thus leading to unpredictable latency for other applications. Thus, we need to enhance the default CUDA MPS to share the GPU judiciously.  The Volta (and newer) generation of NVIDIA GPUs provide a mechanism to set a limit on the amount of GPU resource for each running application by providing a GPU\% as an environmental variable. This fixed GPU\% approach helps us to isolate the GPU resources for a particular application and avoid interference from other applications, thus, guaranteeing predictable latency. TVM fundamentally requires that the profiling server reports the correct time taken by a configuration to execute in the GPU. Therefore, we design a mechanism to share the GPU spatially with fixed GPU\%, while seeking to maximize the GPU utilization. We avoid interference that can occur while sharing the GPU, as seen with temporal sharing or using the default MPS. We discuss the effect of interference in the GPU has on TVM tuning in greater detail in the Appendix.
\newline
We now analyze the impact of tuning a model with one or more TVM servers spatially sharing the GPU using CUDA MPS that is setup to use different fixed GPU\% and its impact on inference latency. 
We also study this interaction across multiple different DNN models. We note that although a DNN tuned at different GPU\% has a different inference latency, the output of the inference \emph{remains the same}. \ie accuracy of the resulting tuned model is not affected.  


\vspace{-2mm}
\subsubsection{Impact on Inference time (Quality of tuned model)} 
\label{sec:impact_on_inference_time}\vspace{-2mm}
We refer the quality of tuned model in terms of the inference time achieved by the tuned model when executed with 100\% GPU and also evaluate the quality based on the variance a model shows in the inference time when executed with different GPU\%.
To measure the impact on inference time we perform inference using the tuned models (tuned at different GPU\%) by explicitly limiting the GPU\% during the inference operation. For this evaluation we  allowed the model to be tuned until the \textit{early stop} option in TVM terminates the tuning. With \textit{early stop}, TVM's schedule explorer checks, and stops tuning, when the new configurations of the convolution operator \Scut{\knote{operator or operation? or is there a DNN term for operator? Is there some word missing before 'operator'? AD: Fixed Please see. Operator is a DNN terminology for a "function" eg. conv2d operator, RELU operator etc.}} do not show latency improvement. 
We use a sample color image of resolution 224$\times$224 pixels to perform inference on. The results are shown in Table~\ref{tab:tuning_percent_vs_inference_percent_1} and Table~\ref{tab:vgg19_tuning_vs_inference}. 
Observe that for all the models, the inference latency is lowest when the GPU\% set for inference matches the GPU\% used while tuning (\ie along the diagonal of the tables). 
We further observe that for the computationally lighter models such as 
ResNet-18 and Mobilenet, 
the models tuned at a higher GPU\% have a relatively larger inference latency when inferred with a lower GPU\% than what it was tuned at. 
\eg A model tuned at 100\% is optimal only when the inference task has 100\% of the GPU, while it has a higher latency when inferred at lower GPU percentages. 
We see this pattern is consistent for both Table~\ref{tab:tuning_percent_vs_inference_percent_1} and Table~\ref{tab:vgg19_tuning_vs_inference}. 
A ResNet-18 model tuned (Table~\ref{tab:resnet-18_tuning_vs_inferece} (left)) at 100\% has 24\% higher latency than the same model tuned at 25\% GPU, when both are inferring with 10\% GPU. 
However, a model tuned at 100\% GPU is only 9\% faster than a model tuned at 25\% GPU, when both are inferred at 100\% GPU. This indicates that there is a "sweet-spot" of GPU\% for tuning a model such that the tuned model provides near-optimal latency over wider range of GPU\% during inference.
However, for computationally denser models such as VGG-19, we do not see as much variation in inference latency (see  
(Table~\ref{tab:vgg19_tuning_vs_inference} for VGG-19). To find the "sweet-spot" we ran an experiment where we take DNN models tuned at different percentages and infer 1000 images (with batch size of 1) each at GPU\% of 10,25,50,75 and 100 and compute the total time taken to infer all 5000 images. We see that for Mobilenet, ResNet-18 and VGG-19, models tuned at 25\% provide the best inference times for 5K images. 

\begin{table}[htbp!]
\vspace{-6mm}
    \centering
    \caption{ResNet-18 \& Mobilenet Inference Latency (ms): tuning and inference at different GPU\%}
    \begin{minipage}{.49\linewidth}
    \label{tab:resnet-18_tuning_vs_inferece}
    \resizebox{\textwidth}{!}{
    \begin{tabular}{c| c c c c c c}
    \toprule
    Inference\% &\multicolumn{5}{c}{Tuning\% (ResNet-18)}\\
    \hline
    {} & 10 & 25 & 50 & 75 & 100 & Untuned \\
    \hline
    10  & {\bf 3.13} & 3.35 & 4.03 & 4.32 & 4.38 & 13.18\\
    25  & 2.06 & {\bf 1.83} & 2.03 & 2.10 & 2.18 & 5.54\\
    50  & 1.75 & 1.41 & {\bf 1.34} & 1.43 & 1.42 & 3.06\\
    75  & 1.71 & 1.27 & 1.21 & {\bf 1.17} & 1.22 & 2.24\\
    100 & 1.64 & 1.22 & 1.17 & 1.11 & {\bf 1.10} & 1.87\\\hline
    \end{tabular}
    }
    \end{minipage}
    \begin{minipage}{.49\linewidth}
    \label{tab:mobilenet_tuning_vs_inference}
    \resizebox{\textwidth}{!}{
    \begin{tabular}{c| c c c c c c}
    \toprule
    Inference\% &\multicolumn{5}{c}{Tuning\% (Mobilenet)}\\
    \hline
    {} & 10 & 25 & 50 & 75 & 100 & Untuned \\\hline
    10  & \bf {1.70} & 1.94 & 2.20 & 2.34 & 2.59 & 2.45\\
    25  & 1.24 & \bf {0.97} & 1.06 & 1.13 & 1.22 & 1.32\\
    50  & 1.13 & 0.77 & \bf {0.67} & 0.71 & 0.73 & 0.79\\
    75  & 1.13 & 0.73 & 0.64 & \bf{0.62} & 0.62 & 0.75\\
    100 & 1.13 & 0.72 & 0.60 & 0.59 & \bf{0.57} & 0.75\\
    \hline
    \end{tabular}
    }
    \end{minipage}
    \label{tab:tuning_percent_vs_inference_percent_1}
    \vspace{-4mm}
\end{table}
\begin{table}[htbp!]
\vspace{-3mm}
    \centering
    \begin{minipage}{.49\linewidth}
   
    \caption{VGG-19 Inference Latency (ms)}
    \resizebox{\textwidth}{!}{
    \begin{tabular}{@{\extracolsep{1pt}}c|cccccc}
    \toprule
    Inference\% &\multicolumn{6}{c}{Tuning\% (VGG-19)}\\
    \hline
    {} & 10 & 25 & 50 & 75 & 100 & Untuned \\
    \hline
    10  & {\bf 15.89} & 16.14 & 16.11 & 16.34 & 16.82 & 16.84\\
    25  & 7.02 & {\bf 6.91} & 7.07 & 7.09 & 7.43 & 7.43\\
    50  & 4.16 & 4.12 & {\bf 4.08} & 4.14 & 4.37 & 4.41\\
    75  & 3.37 & 3.26 & 3.30 & {\bf 3.22} & 3.41 & 3.41\\
    100 & 2.92 & 2.85 & 2.82 &  2.82 & {\bf 2.82} & 3.01\\
    \hline
    \end{tabular}
    }
     \label{tab:vgg19_tuning_vs_inference}
    \end{minipage}
    \begin{minipage}{.49\linewidth}
     \label{tab:total_time_to_infer}
     \caption{Total inference time 5K images (sec.)}
    \resizebox{\textwidth}{!}{
     \begin{tabular}{c| c c c c c}
      \toprule
    Model &\multicolumn{5}{c}{Tuning\%}\\
     \hline
         {}&10&25&50&75&100  \\\hline
         Mobilenet&6.33 & \bf{5.13} & 5.17 & 5.39 & 5.73\\
          ResNet-18&10.29&\bf{9.08}&9.78&10.13&10.3\\
         VGG-19&33.36&\bf{33.28}&33.38&33.61&34.85\\\hline
    \end{tabular}
    
\Scut{
   \begin{tabular}{c| c c c c c c}
   
    \toprule
    Inference\% &\multicolumn{6}{c}{Tuning\% (ResNet-101)}\\
    \hline
    {} & 10 & 25 & 50 & 75 & 100 & Untuned \\
    \hline
    10 & 557& 559 & 564 & 526 & 562 & 565 \\
    25 &226 & 225 &226 &225 &225 &226 \\
    50 &115 &114 & 114 &114 &113 &114 \\
    75 &79 &77 &76 &76 &76 &77 \\
    100 &60 &58 &58 &58 &58 &58\\
    \hline
    \end{tabular}
    }
    }
    \label{tab:total_inference_time}
    \end{minipage}
    \vspace{-4mm}
    \end{table}

\begin{figure}[h!]
\vspace{-4mm}
    \centering
    \includegraphics[width=\linewidth]{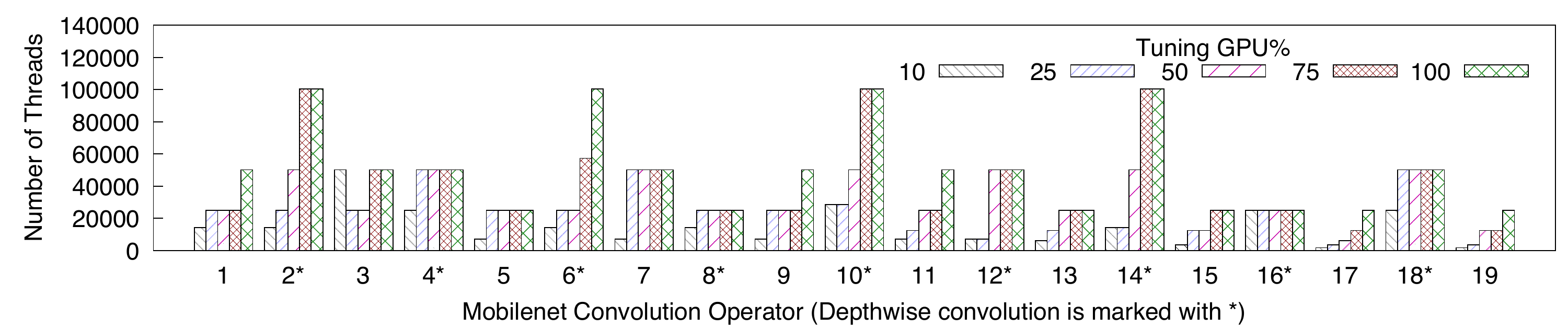}
    \vspace{-6mm}
    \caption{\# of GPU threads in each convolution operation of Mobilenet tuned at different GPU\%} 
    \label{fig:number_of_threads_per_layer}
    \vspace{-4mm}
\end{figure}
\begin{minipage}{\textwidth}
\begin{minipage}{.5\textwidth}
\includegraphics[width=\linewidth]{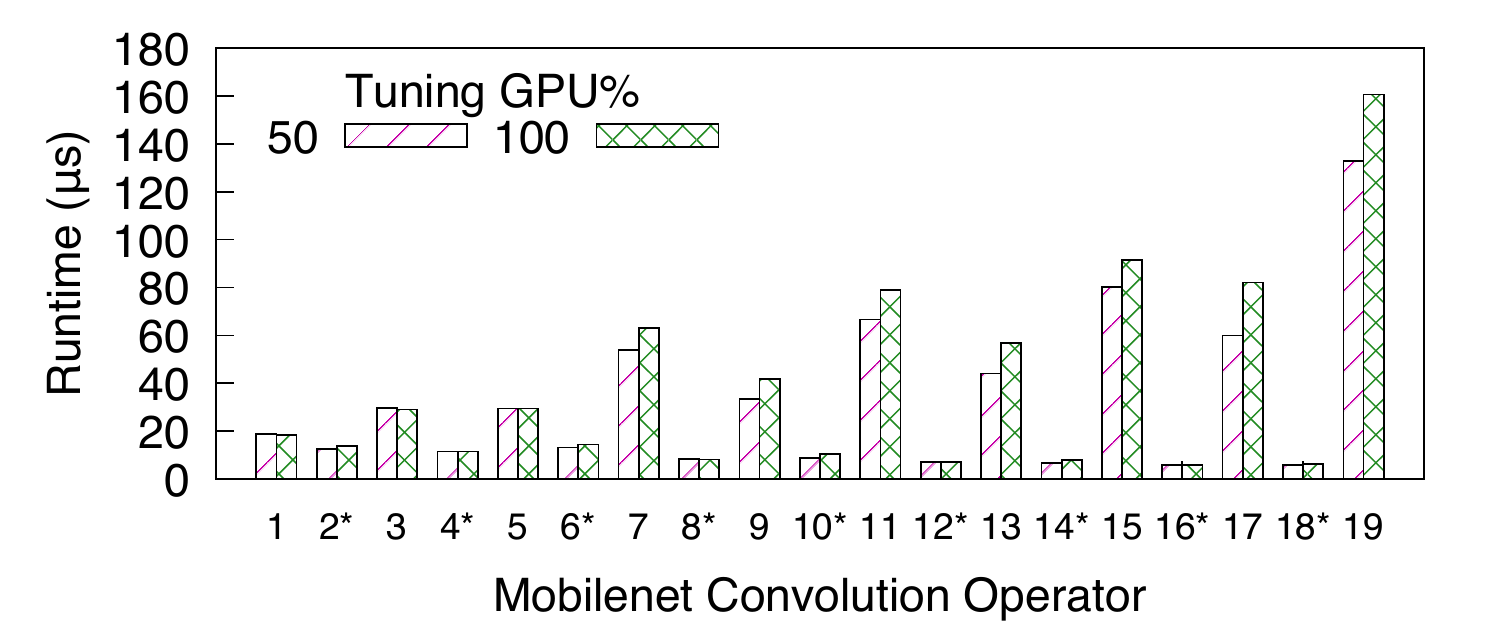}\vspace{-3mm}
\captionof{figure}{
Inference runtime with 25\% GPU. 
}
\label{fig:mobilenet_latency_per_layer}
\end{minipage}
\begin{minipage}{0.5\textwidth}
\vspace{-8mm}
\captionof{table}{Model tuning time (mins) different GPU\%}
\resizebox{\textwidth}{!}{
\begin{tabular}{c| c c c c c}\hline
    Model & \multicolumn{5}{c}{Tuning\%}\\\hline
          & 10\% & 25\% & 50\% & 75\% &100\%   \\\hline
         Mobilenet &628 & 630& 633 & 639 & 630 \\
         ResNet-18 & 498 & 464 & 472 & 466 & 465\\
         VGG-19 & 422 & 408 & 419 & 409  &408\\\hline
    \end{tabular}
}
\label{tab:tuning_time_different_percentage}
\end{minipage}
\end{minipage}

We used the NVIDIA nvprof profiler to profile the tuned models and noted the number of GPU threads each model uses while inferring. This thread count is in Fig.~\ref{fig:number_of_threads_per_layer}. We can see that TVM's tuning picks the configuration with a high thread count for a model tuned at 75\% and 100\% GPU. In an ideal scenario, more threads running concurrently can parallelize the work better, thus achieving lower inference latency. However, 
in a typical GPU only a fixed number of threads (e.g., for V100 GPU, only 2048 GPU threads) can be run in an SM concurrently. While thread count alone does not determine how the SMs in the GPU will be utilized, using more threads does indicate that more SMs are necessary to run those threads concurrently. Thus, if the model with a high thread count is run at lower GPU\%, there will not be enough SMs to run the threads, and hence each operation will take longer to complete.\looseness-1 

To show the impact of having a large number of threads per convolution operations while inferring at low GPU\%, we profiled the runtime of each convolution operation for Mobilenet models tuned at 100\% and 50\%, and then provided 25\% GPU for inference. 
The results are in Figure~\ref{fig:mobilenet_latency_per_layer}.
First, with the model tuned at 100\% GPU, almost all the operators run slower than the model tuned at 50\% GPU. This difference in the runtime is more significant in compute heavy non-depthwise operators (odd numbers in Fig.~\ref{fig:mobilenet_latency_per_layer}). As we noted, the thread count for all the convolution operators in Fig.~\ref{fig:number_of_threads_per_layer} show that the model tuned at 100\% GPU produces operators with a higher number of threads than when it is tuned at 50\%.
Operators with a higher thread count require more GPU resources to run all threads concurrently. Else, some threads have to wait for GPU SMs to free up, thus, increasing the runtime of the operator. Thus, a model tuned at 100\% GPU may be at a disadvantage compared to model tuned at 50\% GPU, when during inference only 25\% GPU is available with fewer available threads. 

\begin{figure}[h!]
\vspace{-5mm}
    \centering
    \includegraphics[width=\linewidth]{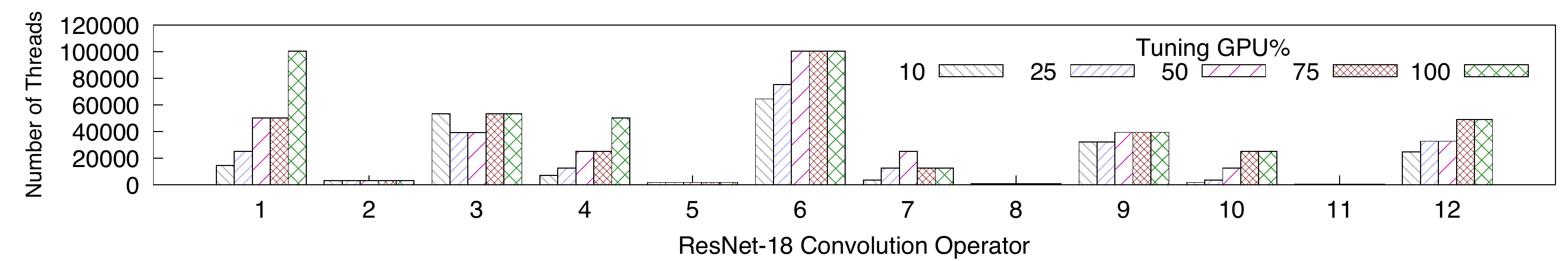}
    \vspace{-6mm}
    \caption{\# of GPU threads used in each ResNet-18 convolution operator (tuned at different GPU\%)}
    \label{fig:number_of_threads_per_layer_resnet}
    \vspace{-3mm}
\end{figure}
We also evaluated 
tuning ResNet-18. We show thread count in  Fig.~\ref{fig:number_of_threads_per_layer_resnet} and note a similar trend as with Mobilenet. 
We should note that for some operators, such as 2,5,8 and 11, the thread count is very low. These are the convolutional operators used in "skip connections" in ResNet and are used to change the dimension of data "skipping" from one layer to another. As they only change the dimension of the matrix, they are relatively light in computation, have lower GFLOPS, and require fewer threads. The remaining convolution operators perform more computation and require a higher number of threads.
\vspace{-2mm}
\subsubsection{Impact on Tuning time}\vspace{-2mm}
We profiled a number of DNN 
models in TVM to understand the impact of varying the GPU\% on total tuning time. 
To have a fair comparison across different GPU\%, we fixed the number of tuning iterations per DNN operator to $1000$.\Scut{ and disabled early stopping mode, so that every convolution layer has to be profiled for the same a fixed 1000 iterations to find the best configuration.} We present the results in Table~\ref{tab:tuning_time_different_percentage}. 
Although the tuning time for a DNN model varies slightly across different GPU\%, the differences are marginal. 
The overall impact on tuning time across all the profiled models is seen to be less than 3\% (the highest variability is for VGG-19). Therefore, we conclude that we can tune a model at a lower GPU\% without adversely affecting the model tuning time.\vspace{-2mm}

%% file: sections/design.tex
\begin{figure}[h]
\vspace{-4mm}
\centering


\begin{subfigure}[b]{0.25\textwidth}
\centering
\includegraphics[width=\textwidth]{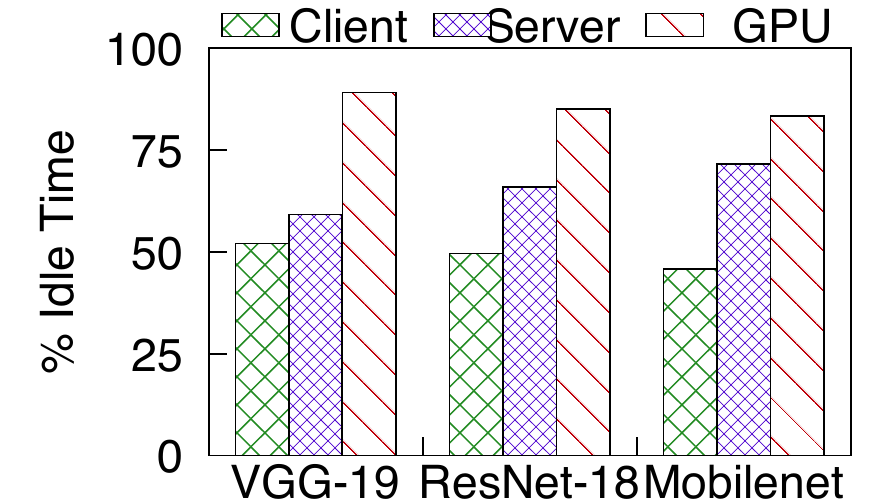}
\caption{Tuning component Idle}
\label{fig:server_idle_time}
\end{subfigure}%
\begin{subfigure}[b]{0.25\textwidth}
\centering
\includegraphics[width=\textwidth]{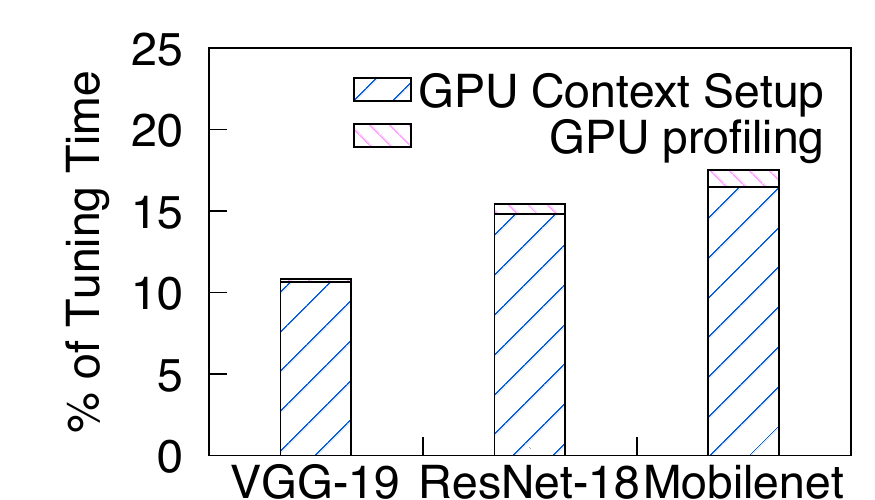}
\caption{GPU Utilization}
\label{fig:gpu_utilization_time}
\end{subfigure}%
\begin{subfigure}[b]{0.20\textwidth}
\centering
\includegraphics[width=\textwidth]{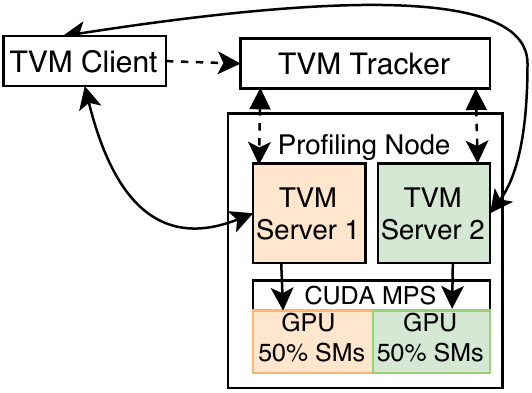}
\caption{2 Servers}
\label{fig:TVM_two_servers}
\end{subfigure}%
\begin{subfigure}[b]{0.25\textwidth}
\centering
\includegraphics[width=\textwidth]{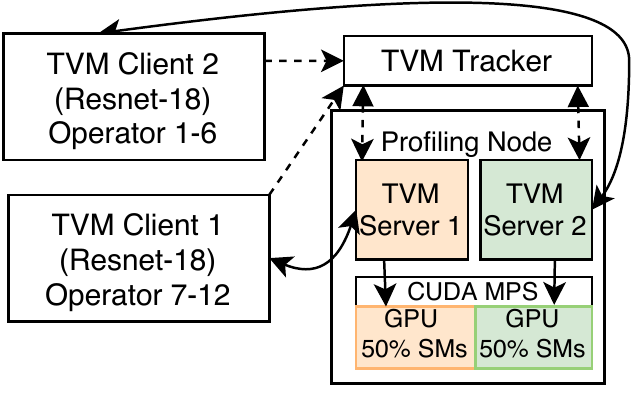}
\caption{Model split on 2 clients}
\label{fig:tvm_split_client}
\end{subfigure}
\label{fig:TVM_Architecture}
\vspace{-0.1cm}
\caption{(a)TVM component idle \%; (b) GPU use breakdown; (c)Server Scaling; (d)Client Sharding \sknote{I think this current overhead plot does not justify the cause. We need to show \% of Idle time first for the CPU and GPU and then the unnecessary overhead (\%) on GPU in the used GPU time.AD: Okay.}}
\vspace{-0.5cm}
\end{figure}

\subsection{Improving the Autotuning Performance and System Utilization}
\vspace{-3mm}
\jnote{I think we need to reorder this section.
I think it will be better to first explain what we can do (e.g., while tuning with less GPU\%, inference latency in deployment increases a little, but we can achieve double throughputs and also do more autotuning tasks at the same time) based on our findings from previous sections and then describe our system (e.g., TSI, and section 3.1.1). So, readers can be convinced why we propose this design and also section 2 and 3 are highly related.}
The default TVM implementation does not recommend using more than one TVM server per GPU device due to interference multiple servers can cause during profiling.
The TVM server receives the configuration files, invokes the GPU for profiling each of the configurations and reports the results back to the client, which then runs an exploration algorithm (simulated annealing~\cite{sa}) and ML algorithm (XGBoost~\cite{chen2016xgboost}) to evaluate what configurations to create for profiling in the next iteration.~\jnote{This is covered in early section like background.} Since the client side processing is also reasonably complex, there is essentially a ping-pong of server-then-client processing for each set of configurations.
This results in very high idle time on the TVM server (and hence the GPU). \sknote{Better to breakdown the Server idle time for both CPU and GPU. Just showing server idle time and hence suggesting GPU also is not correct. Why is the plot that Junguk had in the background section missing in this version? I think that plot was more relevant to motivate and make the case for optimization.} 
\knote{Agreed. If you can also show the GPU idle time on the server, it might be compelling - because the GPU is probably idle 99\% of the time? yes. Please see the plot 8A}\sknote{I think this current overhead plot does not justify the cause. We need to show \% of Idle time first for the CPU and GPU and then the unnecessary overhead (\%) on GPU in the used GPU time.} 
As shown in Fig.~\ref{fig:server_idle_time}, the server remains idle for more than $50\%$ of the total tuning time, while waiting for the client to complete its processing. Furthermore, the GPU is idle for $\sim85\%$ of tuning time. This poor utilization of server resources results in very low tuning throughput\footnote{We define tuning throughput as the number of auto-tuning jobs that can be completed per 1000 minutes.} for the system.
\newline
A common approach to improve utilization 
is to multiplex and concurrently profile multiple (same or different) models. 
For example, concurrently tuning two Resnet-18 models using a single TVM server decreases the server idle time from 50\% to 13\% of the total tuning time. 
However, the overall tuning time of the models increased by about 20\% (\ie from 465 minutes for a single instance to 555 minutes to tune two models of ResNet-18). TVM clients are bursty in nature while profiling configurations,\ie they produce a large number of configurations and profile them at once. This bursty nature causes delay even though the server is still idle for only 13\% of the time.\Scut{\sknote{can we say here the reason for increase? Also what is the mode of GPU multiplexing considered here?}} Nonetheless, the multiplexing approach significantly improves overall tuning throughput ($\sim 40\%$ in this case). \sknote{verify 40\% value} \sknote{Also, can we have numbers for 1 TVM client tuning 1 Resnet-18 model with 2 TVM servers running on the same node -- This is a more natural multiplexing scheme as we are talking about under-utilization in the serer/runner node and nothing about the client side.; This will help make a point for what kind of configuration makes more sense.}

{\bf TVM Server Instance (TSI) Scaling:} 
Tuning (single or multiple models) with multiple TVM server instances on a single server node helps improve resource (CPU and GPU) utilization and also considerably improves overall tuning throughput. However, the cost of multiplexing in terms of the increase in overall tuning time and the side effect on the quality of tuned model \knote{please remember to specify what you mean by quality of tuned model somewhere earlier} and the subsequent impact on inference latency due to the sharing of server resources need to be carefully considered. 
We 
aim to reduce the overall tuning time and increase the GPU utilization by running multiple TSIs concurrently on distinct CPU cores of a multi-core server node in such a way that it does not adversely impact the quality of the tuned model and it does not substantially increase the inference latency. \sknote{I think, this should be our core focus; show adverse impact with other forms of multiplexing and how our approach helps contain the latency impact -- on the same lines we have the motivation to say we may need to run several inference services concurrently and hence spatial or temporal multiplexing such inference instances is inevitable.}
TSI scaling can shard a single model across TSI, with each server instance on a different CPU core on the system. This improves utilization of both CPU and GPU on the server. 
The net effect is increased tuning throughput and lower latency to tune a single model. Additionally, to improve GPU utilization and fully take advantage of the number of CPU cores with multiple TSI running concurrently, we can tune multiple models concurrently and improve overall tuning service throughput. 
Further, we have also observed (Table~\ref{tab:total_inference_time}) that a model tuned with appropriate GPU\% (\eg 25\% for Resnet-18) performs better inference for a wide range of GPU\%.
Thus, we seek to autotune different DNN models with an appropriate GPU\%, which 
also improves GPU utilization.\vspace{-4mm} 

\subsubsection{Scaling auto-tuning performance}
\vspace{-3mm}
Techniques to improve auto-tuning performance, including those described in the previous subsection seek to effectively multiplex multiple TVM server instances (TSI) that profile tasks on the GPU. These techniques utilize spatial sharing of the GPU to provide lower latency and higher throughput while tuning. This spatial sharing provides controlled sharing of the GPU.
\newline\textbf{Spatially sharing with explicit isolation of GPU across multiple TVM servers:}
We launch multiple TSI 
on the same profiling server node and spatially share the GPU by assigning each TSI with a distinct GPU\%. \eg Fig.~\ref{fig:TVM_two_servers} shows two TSIs with 50\% GPU each. \Scut{This approach avoids interference on TSIs as explained in \S~\ref{sec:multiplexing_on_gpu}.} Further, the TVM tracker balances the load from the TVM client equally among different TSIs. Hence, the workload of a single TVM client gets evenly distributed among different TSIs and thus lowers the overall tuning time. 
\newline\textbf{Sharding a model across multiple TVM clients:}
The TVM client (TC) creates new configurations of a DNN operator for profiling that are packaged and sent to the server for profiling. Based on the results obtained from profiling, the subsequent processing at the TC can be substantial (\eg running simulated annealing~\cite{sa} and XGBoost~\cite{chen2016xgboost} for \textit{search strategy} stage and compiling next batch of configuration for target GPU). 
Hence, the TC itself can be a bottleneck while tuning. Also, we observe from Fig.~\ref{fig:server_idle_time}
that the TC processing has the least idle time.
To improve the TC performance, we scale the TC instances (TCI) and shard the convolution operators of the tuning model across different TCIs (that may run on multiple CPU cores). \Scut{\sknote{We need to be consistent in using and differentiating the terms operators, layers,.}} 
This is based on the key insight that auto-tuning a DNN model typically involves tuning distinct layers; but these are often tuned independently and thus can be sharded across multiple TCI.    
When all the TCIs finish tuning their respective layers, we combine the tuning results to get a tuned model. We have verified that model tuned using sharding approach has much lower inference latency 
than an untuned model. 
The final tuned model also has same accuracy as the untuned model, therefore, does not suffer from accuracy loss due to sharding.\looseness-1\\ 
\textbf{System Optimization:} We also performed another key system optimization to improve the TVM tuning time. The default TVM servers fork a child process to carry out the profiling of a configuration. This necessitates every child process having to create a GPU context before profiling the configuration in the GPU. But, the GPU context creation takes about 300 milliseconds, which is a significant portion of average overall time to profile a single configuration (which is \textasciitilde 1.5 sec). GPU context setup accounts for more than 95\% GPU utilization during tuning, while the actual configuration profiling time is very low as seen in Fig.\ref{fig:gpu_utilization_time}. During the initialization time, the TSI process cannot use the GPU to profile the configuration received from the client.
Hence, to avoid the frequent GPU context creation cost, we use a \textbf{long-lived server} which does not fork every new profiling task. Instead, the long lived server profiles the received configuration by executing it as a function. Running a large number of configurations in a single long lived process eliminates significant amount of GPU initialization costs, 
helping to lower the overall model tuning latency.

%% file: sections/evaluation.tex
We evaluate the benefits of our optimizations with the most recent version of the open-source TVM implementation (v0.7) running on a testbed of multicore servers equipped with GPUs in our laboratory. 
We use two identical Dell PowerEdge R720x servers, each with 512 GB of memory and CPU with 40 cores. Each server is equipped with 1 NVIDIA V100 GPU with 16 GB of memory and 80 Streaming Multiprocessors (SMs). Both servers are connected back-back with a 10GbE Ethernet link.   
\vspace{-4mm}
\subsection{Benefits of TSI Scaling and Sharding the model across multiple TCIs}\label{sec:single_client_multiple_server}
\begin{figure}[h]
\vspace{-6mm}
    \centering
    \begin{subfigure}[b]{0.33\textwidth}
    \includegraphics[width=\textwidth]{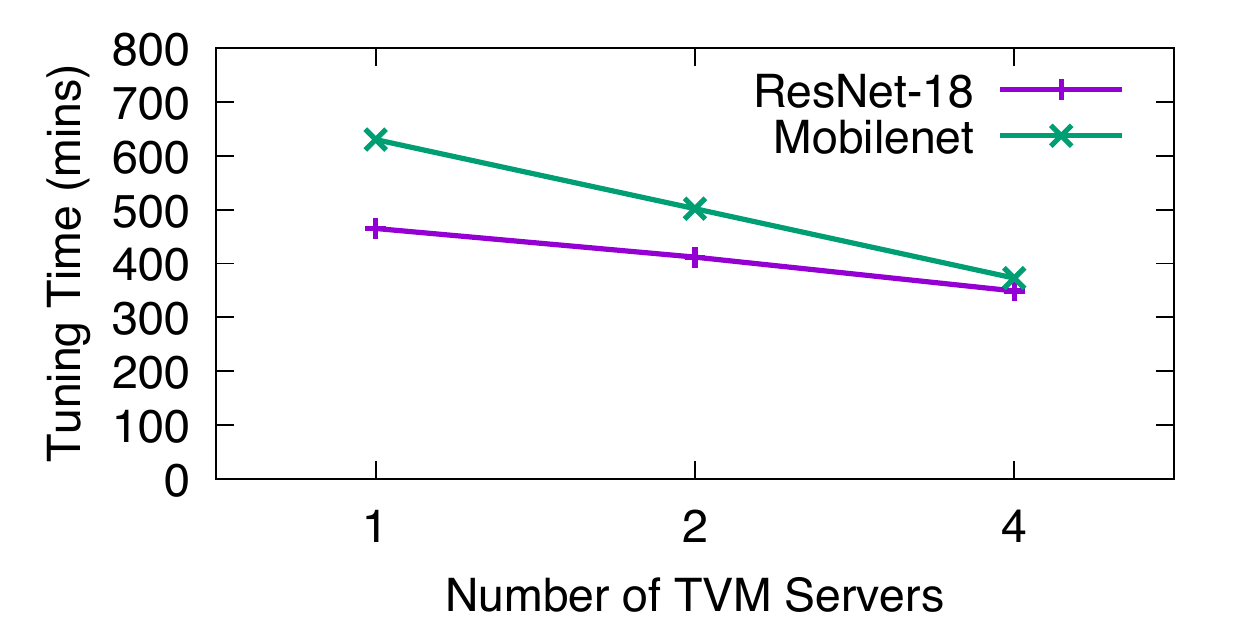}
    \vspace{-4mm}
    \caption{TSI Scaling}
    \label{fig:spatial_sharing_latency}
    \end{subfigure}%
    \begin{subfigure}[b]{0.33\textwidth}
    \includegraphics[width=\textwidth]{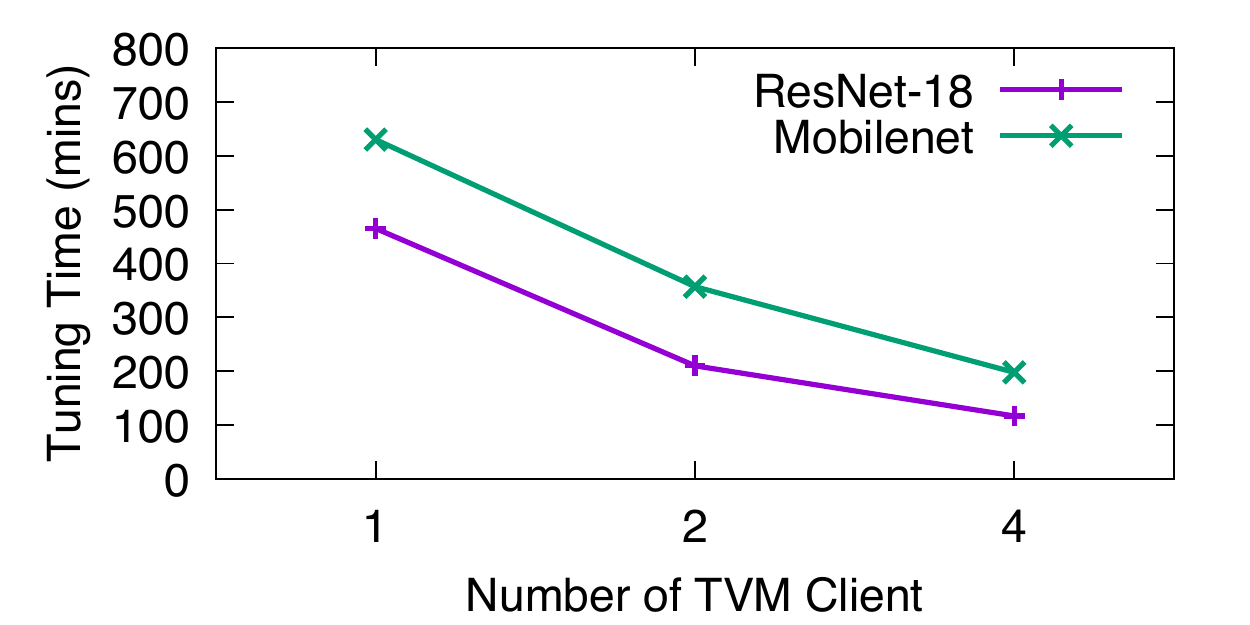}
    \vspace{-4mm}
    \caption{TCI Sharding}
    \label{fig:sharding_model_latency}
    \end{subfigure}%
    \begin{subfigure}[b]{0.33\textwidth}
    \includegraphics[width=\textwidth]{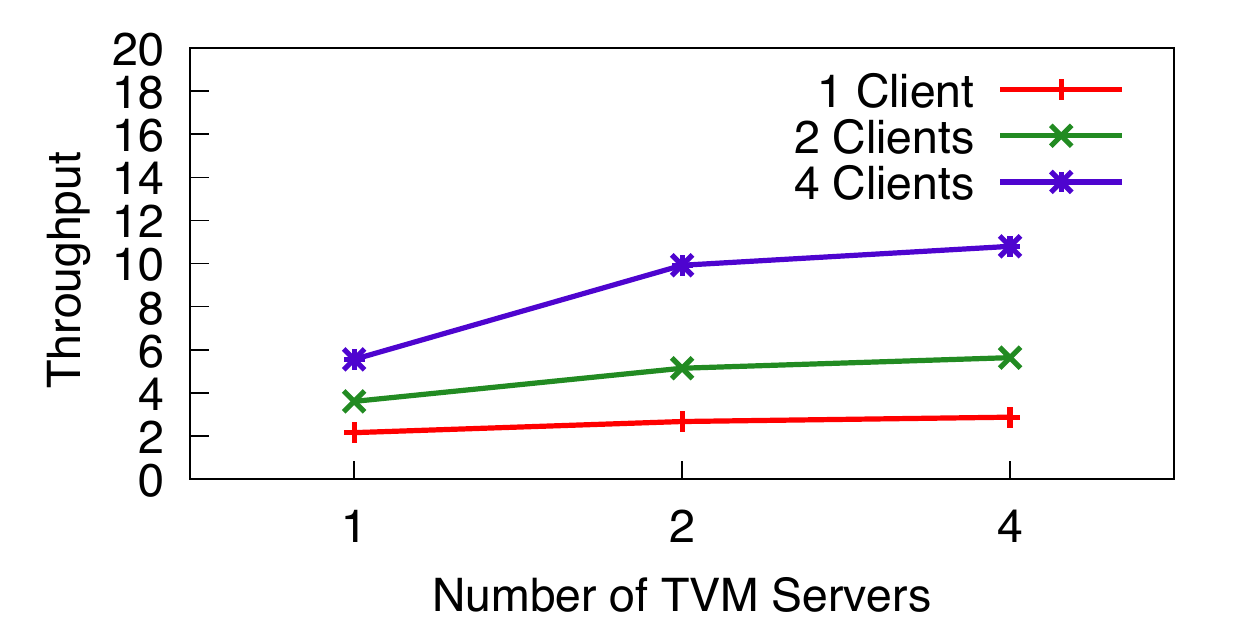}
    \vspace{-4mm}
    \caption{Tput w/Sharding \& Scaling}
    \label{fig:tvm_throughput}
    \end{subfigure}
    \vspace{-3mm}
    \caption{Impact of TSI Scaling and TCI Sharding on Model Tuning Time and Tuning Throughput}
    \vspace{-3mm}
\end{figure}
\vspace{-2mm}
We evaluate the impact on autotuning completion time due to the scaling of TSIs and sharding of the model across multiple TCIs. \sknote{Do we have subserquent improvement factors in the idle time of GPU and CPU resources?}
For these evaluations, we autotuned the ResNet-18 and Mobilenet models in isolation with different TVM client and server instances as distinct experiments. We set the maximum number of tuning iteration to 1000. 
\newline
\textbf{Scaling of TSIs:}
For this experiment, we use a single TC. 
We start with 1 TSI and then scale to 2 and 4 TSIs. \Scut{Further, we also enabled the long-lived server optimization.\sknote{is the long-lived server always on?}}
The results in Fig~\ref{fig:spatial_sharing_latency} show that the tuning time decreases with increasing number of TSIs for both the models. For ResNet-18, scaling to 4 TSIs decreases the tuning time by 25\% (from 465 minutes with single TSI to 349 minutes with 4 TSIs), while Mobilenet shows a 42\% decrease (drops from 630 minutes for 1 TSI to 364 minutes with 4 TSIs).\\ 
\textbf{Sharding the model across multiple TVM clients:} For these experiments we 
shard the models across clients such that the tuning of convolution operators is distributed across multiple TCIs and each TCI uses a single TSI. 
Fig.~\ref{fig:sharding_model_latency} shows that 
sharding helps reduce the tuning time significantly. When compared to tuning a single model with one client, by increasing to 2 TCIs we improve the tuning time by about $48\%$ for ResNet-18 (from 465 minutes to 210 minutes) and $43\%$ for Mobilenet (from 630 minutes to 357 minutes) respectively. With four TCIs, the tuning time further improves from the single tuning instance by 75\% (117 minutes) for ResNet-18 and 68\% (198 minutes) for Mobilenet.\looseness-1\vspace{-3mm}
\vspace{-2mm}
\subsection{Increasing Tuning Throughput}\label{sec:multiple_client_multiple_servers}
\vspace{-3mm}
We evaluated the tuning throughput achieved by our optimization of using multiple TVM profiling servers spatially sharing a GPU. We evaluated the scenario where we have a server node with 1, 2 and 4 profiling TSIs as we vary the number of TCIs (one node each per TCI). For this experiment we use multiple identical ResNet-18 models across all the client instances. 
We present the tuning throughput (models tuned per 1000 minutes) achieved in Figure~\ref{fig:tvm_throughput}. 
\Scut{Increasing the number of clients increases the throughput for a given number of TVM servers. 
This is because TVM servers are idle for a lot of the time during the tuning process. Adding new clients increases multiplexing, and thus the server utilization and throughput.
However, adding more clients to a single TSI increases throughput only marginally as the server (which uses a single CPU core) is close to saturation (max. utilization). 
But, increasing the number of servers (each server has a distinct CPU core) significantly increases throughput.}
Increasing the number of TCIs when there is only one TSI yields a limited increase in throughput, because the single CPU core becomes a limitation. However, with a larger number of TSIs on the server side, 
increasing the number of TCIs then results in significant throughput improvement. 
Adding new TCIs increases multiplexing, and thus the server utilization and throughput. TSIs are no longer underutilized or remain idle for long periods during the tuning process. 
Adding more TSIs and increasing the number of TCIs conflate the benefit by increasing multiplexing, reducing tuning time \emph{and} a substantial increase in throughput.\vspace{-2mm}
\vspace{-2mm}
\subsection{Tuning Multiple Different DNN Models Concurrently}\label{sec:multiple_models_concurrently}\vspace{-2mm}
\begin{minipage}{\textwidth}
\vspace{-0.2cm}
\begin{minipage}[c]{0.35\textwidth}
\centering
   \includegraphics[width=\textwidth]{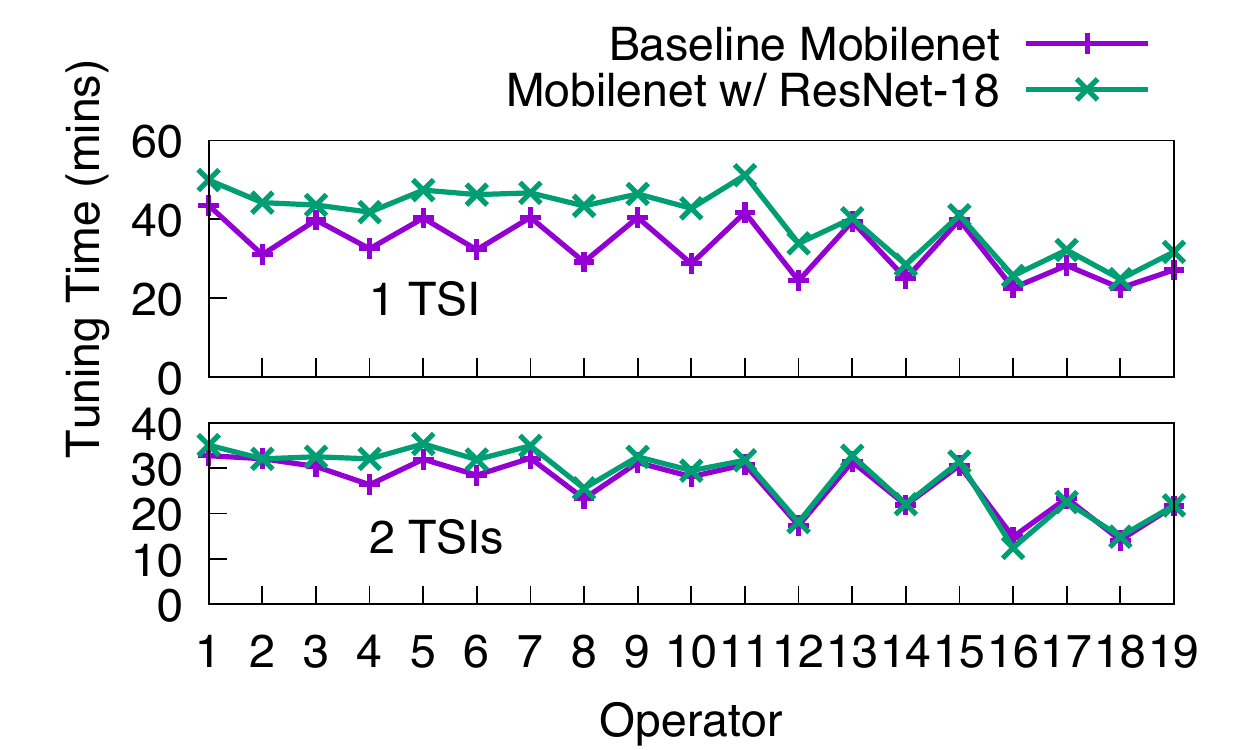}
   \vspace{-6mm}
    \captionof{figure}{Operator Tuning Time}
    \label{fig:two_models_time}
\end{minipage}
\begin{minipage}[c]{0.63\textwidth}
\centering
\includegraphics[width=\linewidth]{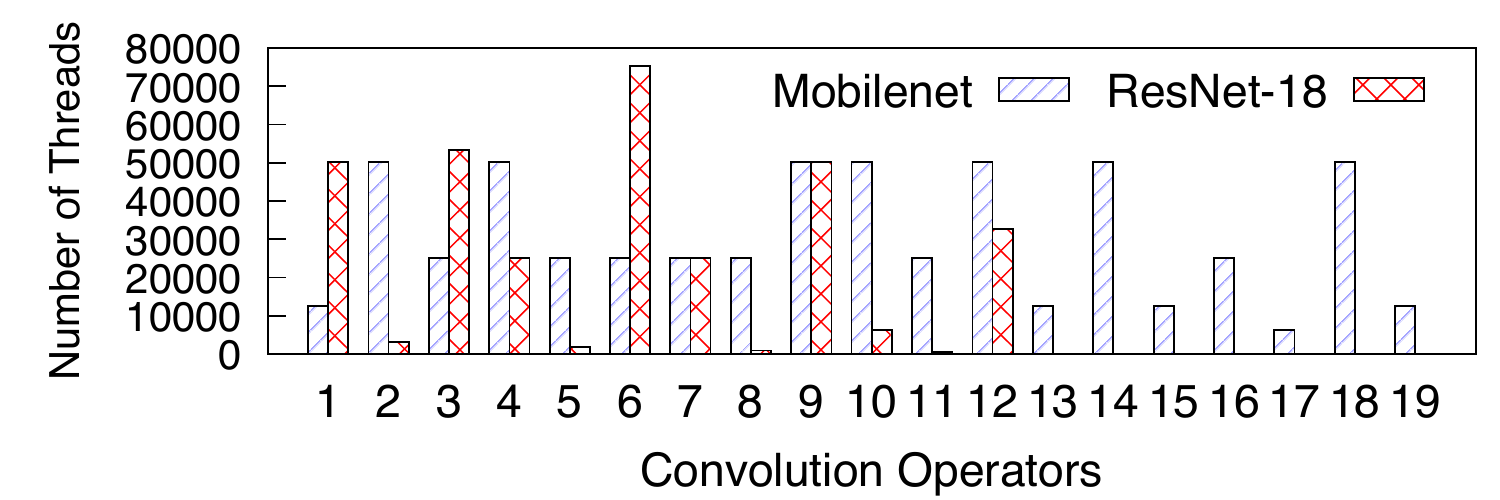}
\vspace{-6mm}
\captionof{figure}{Thread Count of Tuned Models}
\label{fig:mobilenet_final_thread_count}
\Scut{
\captionof{table}{Tuning time \& Thpt with 2 models tuned concurrently}
\resizebox{\textwidth}{!}{\begin{tabular}{c|c c c}\hline
         2 TCI & ResNet-18 & Mobilenet & Throughput\\\hline
         \# of Servers & Time (min.) & Time (min.) & (\# tuned/1000 mins.)\\\hline
         isolated & 465 & 630\\
         1 & 552 & 761 & 1.31\\
         2 & 420 & 519 & 1.92\\
         4 & 340 & 373 & 2.68
    \end{tabular}}
 }
\end{minipage}
\Scut{
\begin{minipage}[c]{0.33\textwidth}
\centering
\captionof{table}{Throughput}
\begin{tabular}{c| c}\hline
         \# of Servers & Throughput\\\hline
         1 & 1.31\\
         2 & 1.92\\
         4 & 2.68\\
    \end{tabular}
    \label{tab:spatial_sharing_different_models}
\end{minipage}
}

\end{minipage}

We now show how combining the scaling of the TSIs for a given model and concurrently having different TCIs for different DNN models can boost a tuning service's throughput through effective multiplexing. We tuned a ResNet-18 and a Mobilenet model using 2 different TCIs, while increasing the number of TSIs from 1 to 4 and evaluated both overall throughput and individual model tuning time as shown in Table~\ref{tab:spatial_sharing_different_models}. 
Having only 1 TSI to concurrently tune the 2 models increases the latency of both models. ResNet-18's tuning time increased by \textasciitilde18\% 
and Mobilenet's by \textasciitilde20\% 
compared to the time taken to tune a single model in isolation. 
But, when we increase the number of TSIs to 2, both models finish tuning within 519 minutes (31\% reduction from the single model in isolation tuning time), and with 4 TSIs, before 373 minutes, a 51\% reduction. Thus, there is improvement for both models once we have an adequate number of TSIs. 
When tuning two different models, one may finish tuning earlier than the other, allowing the slower model to use the additional GPU resources to process, if possible. 
We conservatively estimate the improvement in throughput (models tuned in 1000 minutes) based on the time taken to finish tuning both models. 
\knote{The reader may ask: what if I tuned the 2 models one after the other: 552 plus 761 is 1313. 1000 div. by 656.5 is 1.523. That is the throughput with temporally separately tuning them. So, we have to say that you need to get to 4 TSIs to meaningfully improve throughput.}
Scaling the TSI from 1 to 4 increases essentially doubles the tuning throughput. Compared to separately tuning each model in isolation (throughput of 1.52 models per 1000 mins.), we can tune 2.68 models in the same time by concurrent tuning with an adequate number of TSIs.  
\newline
We also look at the time spent on tuning each operator of Mobilenet (model taking longer to tune) in Fig.~\ref{fig:two_models_time}. After the ResNet-18 model completes tuning at around the 13$^{th}$ operator when using a single TSI (Top plot) and 12$^{th}$ operator when using 2 TSIs (bottom plot) the tuning process speeds up to match the tuning time for baseline Mobilenet. This is because Mobilenet's tuning takes over the TSI and GPU resources freed up by the completion of ResNet-18 tuning.\vspace{-3mm}

\Scut{
\begin{table}[h!]
    \centering
    \caption{Throughput of tuning two different models (ResNet-18 and Mobilenet) with scaling the TVM Servers}
    \begin{tabular}{c| c}\hline
         \# of Servers & Throughput\\\hline
         1 & 1.31\\
         2 & 1.92\\
         4 & 2.68\\
    \end{tabular}
    \Scut{
    \begin{tabular}{c|c c c}\hline
         \# of Servers & ResNet-18 & Mobilenet & Throughput\\\hline
         1 & 552 & 761 & 1.31\\
         2 & 420 & 519 & 1.92\\
         4 & 340 & 373 & 2.68\\
    \end{tabular}
    }
    \label{tab:spatial_sharing_different_models}
\end{table}
}
\vspace{-2mm}
\subsubsection{Quality of Tuned Model}
\vspace{-2mm}
\begin{minipage}{\textwidth}
\begin{minipage}[c]{0.48\textwidth}
\captionof{table}{Tuning time \& Throughput with 2 \\models tuned concurrently}
\vspace{-2mm}
\resizebox{\textwidth}{!}{\begin{tabular}{|c|c c c|}\hline
         2 TCI & ResNet-18 & Mobilenet & Throughput\\\hline
         \# of Servers & Time (min.) & Time (min.) & (\# tuned/1000 mins.)\\\hline
         isolated & 465 & 630 &\\
         1 & 552 & 761 & 1.31\\
         2 & 420 & 519 & 1.92\\
         4 & 340 & 373 & 2.68\\\hline
    \end{tabular}}
     \label{tab:spatial_sharing_different_models}
\end{minipage}
\begin{minipage}[c]{0.29\textwidth}
\centering
\captionof{table}{Long lived \\server tuning time (mins)}
\vspace{-2mm}
\resizebox{\textwidth}{!}{\begin{tabular}{|c| c c|}\hline
         \# TSI & \pbox{20cm}{Default\\ TVM} & \pbox{20cm}{Long-Lived\\ Server}\\\hline
         1   & 465 & 427\\
         2   & 412  &373\\
         4   & 350 & 348\\\hline
\end{tabular}}
\label{tab:long_lived_server}
\end{minipage}
\begin{minipage}[c]{0.20\textwidth}
\centering
\captionof{table}{Chameleon tuning time}
\vspace{-2mm}
\resizebox{\textwidth}{!}{\begin{tabular}{|c| c|}\hline
     \# TSI& \pbox{20cm}{Tuning\\ Time (mins)} \\\hline
     Baseline  & 473\\
     2 & 427\\
     4  & 402\\\hline
\end{tabular}}
\label{tab:chameleon_results}
\end{minipage}
\end{minipage}

We took the ResNet-18 and Mobilenet models tuned in \S\ref{sec:multiple_models_concurrently} using 4 TSIs, each TSI using 25\% GPU. We used those model for inference and profiled the inference with NVIDIA profiler to observe the GPU thread count each operator produces. We present the thread count in Fig.~\ref{fig:mobilenet_final_thread_count}. We can see that most of the operators thread count do not exceed 50,000 mark. We see similar trend in Fig.~\ref{fig:mobilenet_latency_per_layer} and Fig.~\ref{fig:number_of_threads_per_layer_resnet} where the model tuned with 25\% GPU also have most of their operators not exceeding 50,000 threads. Therefore, we see that multiple models tuned concurrently by spatially sharing the GPU produce similar model to one tuned in isolation. 
\Scut{
\begin{figure}[h!]
\centering
   \includegraphics[width=\textwidth]{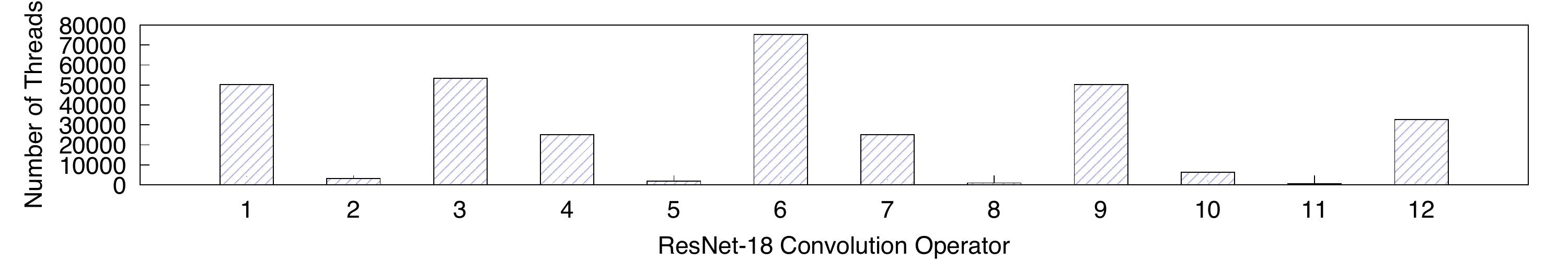}
    \captionof{figure}{ResNet-18 Thread Count}
    \label{fig:resnet-18_final_thread_count}
\end{figure}
}
\vspace{-4mm}
\subsection{Other System Benefits}
\vspace{-2mm}
\textbf{Long-lived Server}:
We also evaluate the impact of having a long-lived server on tuning time of the model. We tune a ResNet-18 model 
using long-lived server, we further increase the number of TSIs. We present our result in Table~\ref{tab:long_lived_server}. We can see that long-lived server helps to drastically reduce the unnecessary overhead on the GPU context creation time and thus reduce the overall tuning time.
\newline\noindent\textbf{Chameleon Autotuning}:
Likewise, we apply the optimization of scaling the TSI to the Chameleon~\cite{chameleon} autotuning platform. We present the results in Table~\ref{tab:chameleon_results}. 
We can observe that TSI scaling decreases the tuning time in Chameleon platform by $\sim 20\%$.
Note: The testbed used for this experiment is different and used a NVIDIA RTX 6000 GPU and 48 Cores CPU node. Hence, the default tuning time is different from the previous experiments we have shown.
\vspace{-3mm}

\Scut{
We further evaluate the quality of tuned model produced by sharing a GPU by observing the latency of the tuned 
\begin{table}[h]
    \centering
    \caption{Inference Latency (ms) (at 100\% GPU) of 4 ResNet-18 Models tuned concurrently with 4 TVM servers sharing GPU temporally, Default MPS and Spatially. Late}
    \begin{tabular}{c|c|c|c|c}\hline
         GPU Sharing & Model 1 & Model 2& Model 3 & Model 4   \\\hline
         Temporal & 1.56 & 1.46 & 1.40 & 1.51\\
         Default MPS & 1.36 & 1.45 & 1.41 & 1.38\\
         Spatial Sharing & 1.20 & 1.12 & 1.11 & 1.15\\
    \end{tabular}
    
    \label{tab:quality_of_model}
\end{table}
}
\Scut{
\subsubsection{Applying \name's Optimization in Chameleon Platform}
Chameleon~\cite{chameleon} is a follow on work, which reduces the TVM's model tuning time by use of Reinforcement learning and clustering. \anote{I will add to this}.
\begin{table}[h]
    \centering
    \begin{tabular}{|c|c|c|c|c|}\hline
         Model Name&Default TVM & Default Chameleon & Chameleon 2 servers & Chameleon 4 Servers  \\\hline
         ResNet-18 &631 & 473 & 427 & 402\\ \hline
    \end{tabular}
    \caption{Time Taken To Tune in Chameleon with \name Optimizations}
    \label{tab:chameleon_results}
\end{table}

}

%% file: sections/related.tex
\textbf{Autotuning systems:} Recent works~\cite{rotem2018glow,chen2018tvm,facebook.tc,liu2019optimizing} propose deep learning compilers to improve the execution efficiency of neural networks (i.e., inference performance) on various hardware.
In addition, extensive efforts~\cite{chen2018learning,ahn2019reinforcement,chameleon,tomczak2019simulating} have been made to address performance problems (i.e., long autotuning completion time) in autotuning systems by enhancing exploration algorithms~\cite{chen2018learning,ahn2019reinforcement,chameleon} and ML cost models~\cite{chen2018learning,tomczak2019simulating}.
Our work is complementary to those works, but differs from them in the sense that we propose a system to reduce autotuning time by multiplexing resources (e.g., GPU) for autotuning procedures and optimizing a profiling server (e.g,. avoiding heavy CUDA initialization overhead).
\newline
\textbf{GPU sharing for ML Inference:} Recent works~\cite{jain2018dynamic,romero2019infaas}
have shown sharing GPU resources for inference help to improve GPU utilization. Our work differs from them. First, we focus on sharing GPU resources for autotuning procedures. Second, we studied GPU sharing impact on inference performance based on tuned models from the autotuning procedures.

%% file: sections/summary.tex
In this paper, we make the case for controlled spatial sharing of GPU for tuning DNN models.
We noted the inter-dependency between model tuning  and inference with the tuned model with an  appropriate GPU\%.
DNN models tuned with the appropriate GPU resources (GPU\%) provide better models, having lower inference latency over a wider range of GPU\%, than models tuned with 100\% GPU. We also show that tuning with a high GPU\% results in a model that requires a large number of threads for each convolution operator, thus incurring additional latency during inference, especially when inference is done with less hardware resources (lower GPU\%). 
Based on these observations, we recommend the DNN models to be tuned at an appropriate GPU\% (several models we tuned needed no more than 25\% GPU) instead of 100\%, which also allows for better multiplexing and utilize the remaining GPU resources to tune other DNN models. 
We present the mechanisms namely, TSI scaling, TCI sharding the model and having a long-lived server process that reduce tuning time by up to 75\% and achieve up to 5-fold increase in tuning throughput.

%% file: neurips_2020.bbl
\begin{thebibliography}{10}

\bibitem{nvidia.mps}
{Multi-Process Service}.
\newblock \url{https://docs.nvidia.com/deploy/mps/index.html}, 2019.

\bibitem{ahn2019reinforcement}
Byung~Hoon Ahn, Prannoy Pilligundla, and Hadi Esmaeilzadeh.
\newblock Reinforcement learning and adaptive sampling for optimized dnn
  compilation.
\newblock {\em arXiv preprint arXiv:1905.12799}, 2019.

\bibitem{chameleon}
Byung~Hoon Ahn, Prannoy Pilligundla, and Hadi Esmaeilzadeh.
\newblock Chameleon: Adaptive code optimization for expedited deep neural
  network compilation.
\newblock {\em ICLR 2020}, 2020.

\bibitem{chen2016xgboost}
Tianqi Chen and Carlos Guestrin.
\newblock Xgboost: A scalable tree boosting system.
\newblock In {\em Proceedings of the 22nd acm sigkdd international conference
  on knowledge discovery and data mining}, pages 785--794, 2016.

\bibitem{chen2018tvm}
Tianqi Chen, Thierry Moreau, Ziheng Jiang, Lianmin Zheng, Eddie Yan, Haichen
  Shen, Meghan Cowan, Leyuan Wang, Yuwei Hu, Luis Ceze, et~al.
\newblock $\{$TVM$\}$: An automated end-to-end optimizing compiler for deep
  learning.
\newblock In {\em 13th $\{$USENIX$\}$ Symposium on Operating Systems Design and
  Implementation ($\{$OSDI$\}$ 18)}, pages 578--594, 2018.

\bibitem{chen2018learning}
Tianqi Chen, Lianmin Zheng, Eddie Yan, Ziheng Jiang, Thierry Moreau, Luis Ceze,
  Carlos Guestrin, and Arvind Krishnamurthy.
\newblock Learning to optimize tensor programs.
\newblock In {\em Advances in Neural Information Processing Systems}, pages
  3389--3400, 2018.

\bibitem{isscc_2016_chen_eyeriss}
{Chen, Yu-Hsin and Krishna, Tushar and Emer, Joel and Sze, Vivienne}.
\newblock {Eyeriss: An Energy-Efficient Reconfigurable Accelerator for Deep
  Convolutional Neural Networks}.
\newblock In {\em {IEEE International Solid-State Circuits Conference, ISSCC
  2016, Digest of Technical Papers}}, pages {262--263}, {2016}.

\bibitem{dhakal2019netml}
Aditya Dhakal and K.~K. Ramakrishnan.
\newblock Netml: An nfv platform with efficient support for machine learning
  applications.
\newblock In {\em 2019 IEEE Conference on Network Softwarization (NetSoft)},
  pages 396--404. IEEE, 2019.

\bibitem{fc.ml.2018}
Kim Hazelwood, Sarah Bird, David Brooks, Soumith Chintala, Utku Diril, Dmytro
  Dzhulgakov, Mohamed Fawzy, Bill Jia, Yangqing Jia, Aditya Kalro, et~al.
\newblock Applied machine learning at facebook: A datacenter infrastructure
  perspective.
\newblock In {\em 2018 IEEE International Symposium on High Performance
  Computer Architecture (HPCA)}, pages 620--629. IEEE, 2018.

\bibitem{hill2017deftnn}
Parker Hill, Animesh Jain, Mason Hill, Babak Zamirai, Chang-Hong Hsu, Michael~A
  Laurenzano, Scott Mahlke, Lingjia Tang, and Jason Mars.
\newblock Deftnn: Addressing bottlenecks for dnn execution on gpus via synapse
  vector elimination and near-compute data fission.
\newblock In {\em Proceedings of the 50th Annual IEEE/ACM International
  Symposium on Microarchitecture}, pages 786--799, 2017.

\bibitem{jain2018dynamic}
Paras Jain, Xiangxi Mo, Ajay Jain, Harikaran Subbaraj, Rehan~Sohail Durrani,
  Alexey Tumanov, Joseph Gonzalez, and Ion Stoica.
\newblock Dynamic space-time scheduling for gpu inference.
\newblock {\em arXiv preprint arXiv:1901.00041}, 2018.

\bibitem{jeon2019analysis}
Myeongjae Jeon, Shivaram Venkataraman, Amar Phanishayee, Junjie Qian, Wencong
  Xiao, and Fan Yang.
\newblock Analysis of large-scale multi-tenant $\{$GPU$\}$ clusters for
  $\{$DNN$\}$ training workloads.
\newblock In {\em 2019 $\{$USENIX$\}$ Annual Technical Conference
  ($\{$USENIX$\}$$\{$ATC$\}$ 19)}, pages 947--960, 2019.

\bibitem{jouppi2017datacenter}
Norman~P Jouppi, Cliff Young, Nishant Patil, David Patterson, Gaurav Agrawal,
  Raminder Bajwa, Sarah Bates, Suresh Bhatia, Nan Boden, Al~Borchers, et~al.
\newblock In-datacenter performance analysis of a tensor processing unit.
\newblock In {\em Proceedings of the 44th Annual International Symposium on
  Computer Architecture}, pages 1--12, 2017.

\bibitem{sa}
Scott Kirkpatrick, C~Daniel Gelatt, and Mario~P Vecchi.
\newblock Optimization by simulated annealing.
\newblock {\em science}, 220(4598):671--680, 1983.

\bibitem{liu2019optimizing}
Yizhi Liu, Yao Wang, Ruofei Yu, Mu~Li, Vin Sharma, and Yida Wang.
\newblock Optimizing $\{$CNN$\}$ model inference on cpus.
\newblock In {\em 2019 $\{$USENIX$\}$ Annual Technical Conference
  ($\{$USENIX$\}$$\{$ATC$\}$ 19)}, pages 1025--1040, 2019.

\bibitem{DBLP:journals/corr/MolchanovTKAK16}
Pavlo Molchanov, Stephen Tyree, Tero Karras, Timo Aila, and Jan Kautz.
\newblock Pruning convolutional neural networks for resource efficient transfer
  learning.
\newblock {\em CoRR}, abs/1611.06440, 2016.

\bibitem{romero2019infaas}
Francisco Romero, Qian Li, Neeraja~J Yadwadkar, and Christos Kozyrakis.
\newblock Infaas: Managed \& model-less inference serving.
\newblock {\em arXiv preprint arXiv:1905.13348}, 2019.

\bibitem{rotem2018glow}
Nadav Rotem, Jordan Fix, Saleem Abdulrasool, Garret Catron, Summer Deng, Roman
  Dzhabarov, Nick Gibson, James Hegeman, Meghan Lele, Roman Levenstein, et~al.
\newblock Glow: Graph lowering compiler techniques for neural networks.
\newblock {\em arXiv preprint arXiv:1805.00907}, 2018.

\bibitem{tomczak2019simulating}
Jakub~M Tomczak, Romain Lepert, and Auke Wiggers.
\newblock Simulating execution time of tensor programs using graph neural
  networks.
\newblock {\em arXiv preprint arXiv:1904.11876}, 2019.

\bibitem{facebook.tc}
Nicolas Vasilache, Oleksandr Zinenko, Theodoros Theodoridis, Priya Goyal,
  Zachary DeVito, William~S Moses, Sven Verdoolaege, Andrew Adams, and Albert
  Cohen.
\newblock Tensor comprehensions: Framework-agnostic high-performance machine
  learning abstractions.
\newblock {\em arXiv preprint arXiv:1802.04730}, 2018.

\end{thebibliography}
